# DeepRGVP: A novel microstructure-informed supervised contrastive learning framework for automated identification of the retinogeniculate pathway using dMRI tractography

Sipei Li[1,2], Jianzhong He[2,5], Tengfei Xue[2,6], Guoqiang Xie[2,3], Shun Yao[2,4], Yuqian Chen[2,6], Erickson F. Torio[2], Yuanjing Feng[5], Dhiego CA Bastos[2], Yogesh Rathi[2], Nikos Makris[2], Ron Kikinis[2], Wenya Linda Bi[2], Alexandra J Golby[2], Lauren J O'Donnell[2]\*, Fan Zhang[2]\*

[1] University of Electronic Science and Technology of China, Chengdu, China
[2] Harvard Medical School, Boston, USA
[3] Nuclear Industry 215 Hospital of Shaanxi Province, Xianyang, China
[4] The First Affiliated Hospital, Sun Yat-sen University, Guangzhou, China
[5] Zhejiang University of Technology, Hangzhou, China
[6] University of Sydney, Sydney, Australia

## ABSTRACT

The retinogeniculate pathway (RGVP) is responsible for carrying visual information from the retina to the lateral geniculate nucleus. Identification and visualization of the RGVP are important in studying the anatomy of the visual system and can inform treatment of related brain diseases. Diffusion MRI (dMRI) tractography is an advanced imaging method that uniquely enables *in vivo* mapping of the 3D trajectory of the RGVP. Currently, identification of the RGVP from tractography data relies on expert (manual) selection of tractography streamlines, which is time-consuming, has high clinical and expert labor costs, and affected by inter-observer variability. In this paper, we present what we believe is the first deep learning framework, namely *DeepRGVP*, to enable fast and accurate identification of the RGVP from dMRI tractography data. We design a novel microstructure-informed supervised contrastive learning method that leverages both streamline label and tissue microstructure information to determine positive and negative pairs. We propose a simple and successful streamline-level data augmentation method to address highly imbalanced training data, where the number of RGVP streamlines is much lower than that of non-RGVP streamlines. We perform comparisons with several state-of-the-art deep learning methods that were designed for tractography parcellation, and we show superior RGVP identification results using DeepRGVP.

## 1. INTRODUCTION

The retinogeniculate pathway (RGVP) is responsible for carrying visual information from the retina to the lateral geniculate nucleus (LGN) [1,2]. It consists of three anatomical segments, including the optic nerve, the optic chiasm and the optic tract [1]. The RGVP is affected in many diseases, including pituitary tumors [3], optic neuritis [4], optic nerve sheath meningiomas [5] and many others [6–8]. Identification and visualization of the RGVP are important in studying the anatomy of the visual system [9] and can inform treatment of brain diseases such as lesions intrinsic or extrinsic to the pathway [10].

Diffusion MRI (dMRI) tractography is an advanced imaging method [11] that uniquely enables *in vivo* reconstruction of the 3D streamline trajectory of the RGVP in a non-invasive way. Many tractography-based studies have shown successful mapping of the RGVP for clinical and research purposes [12–16]. Currently, identification of the RGVP from tractography relies on expert selection, where streamlines are selected based on whether they end in and/or pass through regions of interest (ROIs) drawn by experts [17]. ROI-based RGVP selection, however, is time-consuming, is inefficient with high clinical and expert labor costs, and is also affected by inter-observer variability depending on the experience of experts. Therefore, there is a high need for computational methods to enable automated identification of the RGVP.

In recent years, deep-learning-based methods have been demonstrated to be successful in tractography parcellation for automated identification of white matter fiber tracts in the cerebrum [18–21]. To the best of our knowledge, there are no automated methods designed for RGVP identification yet. Recent advances in deep learning provide a promising approach to enable accurate and fast identification of the RGVP. However, there are two key challenges, as follows.

First, current deep-learning-based tractography parcellation methods mostly use streamline geometric features extracted based on the streamline point spatial coordinates [20,22–24], e.g., RAS (Right, Anterior, Superior) coordinates. While this spatial information is effective to differentiate white matter fiber tracts that have large shape and position dissimilarities, the geometric differences between RGVP streamlines and the nearby non-RGVP streamlines can be very small (see Fig. 1(a)). Thus, only using geometric features can be ineffective for RGVP identification. In tractography data, in addition to

---

streamline geometric features, other informative features can be computed, including diffusion microstructure measures, e.g., the widely used fractional anisotropy (FA). We hypothesize that including such microstructure features can improve RGVP identification. Our rationale is that streamlines representing similar anatomical structures should not only have a similar geometric trajectory but also should have a very similar FA value.

Second, in tractography data there can be a highly imbalanced streamline sample distribution between RGVP and non-RGVP streamlines (our data shows a ratio of 1:8 between RGVP and non-RGVP streamlines; see Sec. 2.1). Imbalanced training data is a well-known challenge in deep learning and limits network generalization for small-size sample categories [25]. Data augmentation (DA) is one effective solution to resolve this challenge. In related work, several tractography studies have performed DA by generating synthetic data samples, e.g., repeating and/or adding noise to the existing tractography data [22,26]. One recent study has proposed a subject-level tractography DA strategy that generates multiple new datasets by downsampling each subject's tractography data, without creating synthetic data [27]. Inspired by this work, we propose a new streamline-level DA method that uses random subsampling to increase the sample size of RGVP streamlines. We hypothesize that this method can generate a balanced dataset for improved model training.

In light of the above, this study presents a novel deep learning framework, namely *DeepRGVP*, for automated identification of the RGVP using dMRI tractography. DeepRGVP is based on the Superficial White Matter Analysis (SupWMA) method [24], a point-cloud-based network [28] with supervised contrastive learning (SCL) [29], which is designed for classification of superficial white matter streamlines. In this study, our contributions are as follows. First, we present what we believe is the first deep learning approach that enables fast and accurate RGVP identification. Second, we design a microstructure-informed SCL (MicroSCL) method that leverages both streamline label (RGVP and non-RGVP) and tissue microstructure (FA) information to determine positive and negative pairs. Third, we design a simple and successful streamline-level data augmentation (StreamDA) method to address the imbalanced training data problem. Compared to several state-of-the-art (SOTA) methods that were designed for tractography parcellation, we demonstrate superior RGVP identification results using DeepRGVP.

## 2. METHODOLOGY

Our overall goal is to identify streamlines that belong to the RGVP from input tractography data generated in the skull base region, as overviewed in Fig 1. Fig. 1(a) gives a visualization of example input tractography data, RGVP streamlines selected using expert-drawn ROIs, and all other unselected non-RGVP streamlines (see Sec. 2.1). The RGVP and non-RGVP streamlines are visually highly similar in terms of their geometric trajectory; however, the non-RGVP streamlines fail to satisfy the strict anatomical ROI selection criteria. We can also observe different FA values in local streamline regions of RGVP and non-RGVP streamlines (indicated using red arrows). This motivates the design of our MicroSCL method (Sec. 2.2.1). Fig. 1(b) demonstrates our StreamDA method (Sec. 2.2.2) to resolve any potential training biases due to the highly imbalanced input data. Additional streamline samples are generated for each RGVP streamline so that a balanced training dataset is achieved. Fig. 1(c) shows the overall network architecture that includes a SCL subnetwork to learn the global feature for each input streamline and a downstream subnetwork to classify the streamlines into RGVP and non-RGVP.

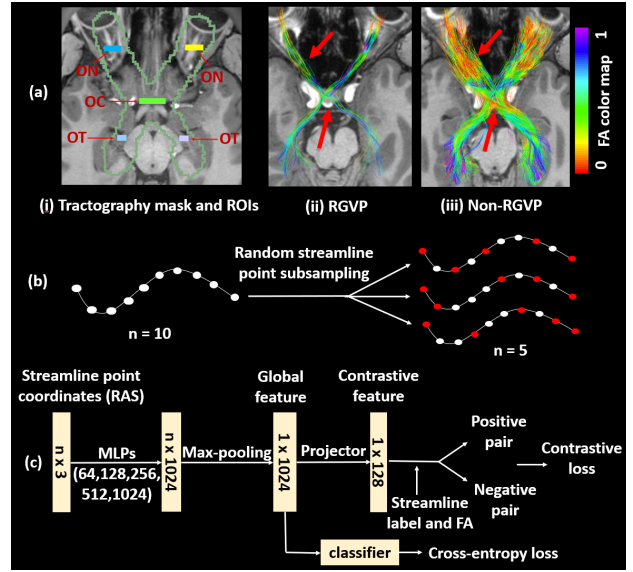

Fig. 1. Overview of DeepRGVP. (a) Example input tractography data and ROIs used for generating ground truth data. (b) Graphic illustration of the StreamDA method. (c) The overall network architecture, including the MicroSCL network and the downstream classifier for streamline classification.

### 2.1 dMRI data, tractography and ground truth

We use a total of 100 dMRI datasets from the Human Connectome Project (HCP) database [30]. The HCP data was acquired with a high-quality image acquisition protocol and preprocessed for data artifact correction [31]. The acquisition parameters are: TE/TR=89.5/5520 ms, and voxel size=1.25x1.25x1.25 mm$^3$. We use the single-shell b=1000 s/mm$^2$ data because it has been shown to enable highly effective RGVP tracking [1]. We perform a visual check of the dMRI data for each of the 100 subjects and exclude 38 subjects that had dMRI data with incomplete RGVP coverage due to face removal for data anonymization, as described in our previous work [1]. Thus, tractography data from 62 subjects are used.

Tractography is performed with a mask (Fig. 1(a.i)) in using two-tensor unscented Kalman filter (UKF) tractography [32] via SlicerDMRI [33,34]. We choose UKF because it can accurately track the RGVP [1] and other cranial nerves [35,36] and it allows estimation of streamline-specific microstructure measures, including the measure of interest, FA. A streamline length threshold of 80 mm (a value lower than the length ~100 mm of ground truth RGVP streamlines) is applied to eliminate any effect from streamlines too short to form the RGVP.

We leverage ground truth RGVP streamlines (Fig. 1(a.ii)) selected using ROIs drawn by an expert (practicing neurosurgeon G.X.) and multi-rater validated in our previous study [1]. These ROIs include the optic nerve (ON), optic chiasm (OC), and optic tract (OT) (Fig. 1(a.i)). For model training, we also leverage the non-RGVP streamlines (Fig. 1(a.iii)) not selected by these ROIs. On average, there are ~150 RGVP and ~1200 non-RGVP streamlines per subject (thus, a ratio of ~1:8 of RVGP to non-RGVP samples for training).

## 2.2 Deep learning framework for RGVP identification

DeepRGVP is based on a SOTA deep learning framework, SupWMA [24], that was developed for parcellation of superficial white matter in the brain. SupWMA can leverage input positive and negative streamline samples, incorporates SCL based on streamline labels, and benefits from an input feature representation as point clouds [24]. It includes an encoder that extracts the global feature for each streamline, and a classifier that predicts a streamline label. DeepRGVP extends SupWMA with two innovative additions: 1) a microstructure-informed SCL (MicroSCL) method, and 2) a streamline-level data augmentation (StreamDA) method, as described below.

*2.2.1 Microstructure-informed supervised contrastive learning: MicroSCL*
In addition to the traditional usage of label information (RGVP and non-RGVP) for positive and negative sample pair determination [29], we propose to include additional information about tissue microstructure to improve pair determination. In this way, the learned global feature may better differentiate streamlines with similar trajectories but from different classes. To achieve this, we compute the absolute difference of mean streamline FA between each streamline pair. We constrain positive pairs to be streamlines from the same class that satisfy the following:

$$\Delta FA = |FA_i - FA_p| < T_{FA} \quad (1)$$

where $T_{FA}$ is a threshold on the allowable FA difference between streamlines in a positive pair.

Overall, the supervised contrastive loss $L_{MicroSCL}$ used in our study is:

$$L_{MicroSCL} = \sum_{i \in I} \frac{-1}{|P(i)|} \sum_{i \in P(i)} log \frac{exp(z_i \cdot z_p/\tau)}{\sum_{i \in A(i)} exp(z_i \cdot z_a/\tau)}$$
$$= \sum_{i \in I} \frac{-1}{|M(i) \cap N(i)|} \sum_{i \in M(i) \cap N(i)} log \frac{exp(z_i \cdot z_{m \cap n}/\tau)}{\sum_{i \in A(i)} exp(z_i \cdot z_a/\tau)} \quad (2)$$

where $i$ is a streamline belonging to a training batch $I$; $M(i)$ is the streamline set of the same class label as streamline $i$; $N(i)$ is the streamline set that satisfies the $\Delta FA$ condition in Eq (1); $P(i)$ is the intersection of $M(i)$ and $N(i)$; $A(i)$ is the set that includes all streamlines except for streamline $i$ in batch $I$; $z_i$, $z_p$ and $z_a$ are contrastive features of streamlines $i$, $p \in P(i)$ and $a \in A(i)$, respectively; $\tau$ (temperature) is a hyperparameter for optimization predefined to be 0.1 as suggested in [37].

In the downstream subnetwork, the classifier predicts streamline class according to the global feature generated from MicroSCL (Fig. 1(c)). A simple network including 3 fully connected (FC) layers with sizes of 512, 256, and 2 (number of classes) and a cross-entropy loss is used.

*2.2.2 Streamline-level data augmentation: StreamDA*
In order to curate a training dataset with a balanced sample distribution for improved model training, we propose the following StreamDA method. Each streamline consists of a sequence of points estimated by a tractography algorithm. For input to the network, we represent each streamline using $P$ points sampled along the streamline. For DA, we generate additional samples from each streamline by repeating the streamline point subsampling process multiple times, such that each time a different point subset is generated (as demonstrated in Fig 1(b)). We note that StreamDA is different from commonly used DA strategies, such as data repetition and adding noise, where additional samples are synthetically generated. In our study, given the original 1:8 ratio of RGVP to non-RGVP streamlines, we perform the StreamDA process 8 times for each RGVP streamline.

For better model training with the augmented data, we also modify the SupWMA network with additional layers. In SupWMA, there are 3 shared multi-layer perceptron (MLP) layers with sizes 64, 128 and 1024 in the encoder network. In the current study, we add two additional MLP layers with sizes 256 and 512 after the layer of size 128, as shown in Fig. 1(c).

## 2.3 Implementation details

Our method is implemented using Pytorch (v1.7) [38] and model training is performed on a NVIDIA GeForce GTX 1080 Ti machine. All hyperparameters are set to the default settings suggested in SupWMA, except for a modified learning rate (0.01) and batch size (512) that are tuned to accompany the addition of FA for pair determination. The threshold $T_{FA}$ in Eq (2) is set to be 0.1 (parameter search from 0.01 to 0.5), and the number of points $P$ is set to 60 (performance increases as $P$ increases but also increases the computational burden; thus, as a trade-off, we choose $P$=60). Both training phases utilize Adam [39] as the optimizer with no weight decay. On average, each training epoch takes 4 seconds with 3GB GPU memory usage when using StreamDA. The code will be made available upon request.

## 3. EXPERIMENTAL RESULTS AND DISCUSSION

We perform evaluation using data from 62 HCP subjects (Sec 2.1), including 40 subjects for training, 10 subjects for validation, and 12 subjects for testing. We compare DeepRGVP with several SOTA deep learning methods that were designed for tractography parcellation. We also perform an ablation study to demonstrate the performance of DeepRGVP's sub components including the MicroSCL and StreamDA methods. The evaluation metrics include precision, recall, F1 and classification accuracy.

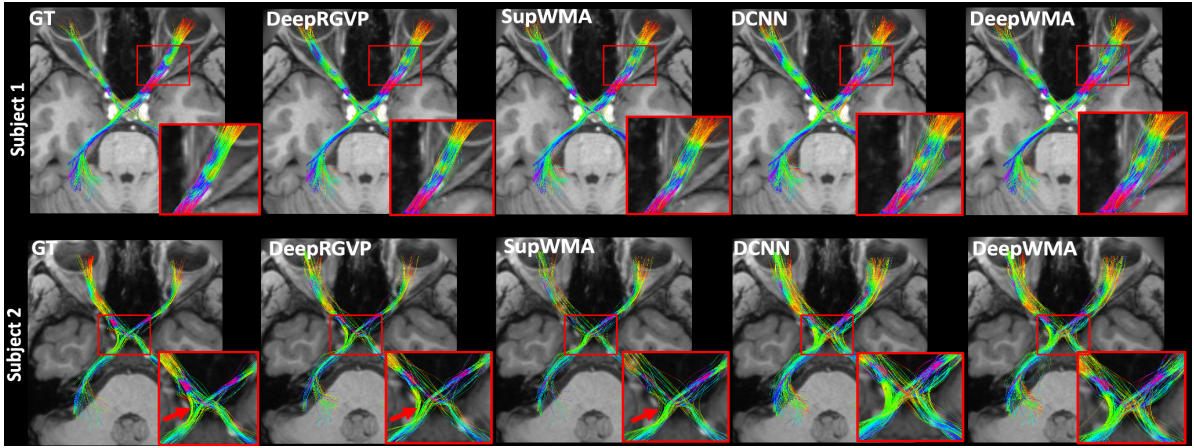

Fig. 2. Visual comparison of the identified RGVPs in two example subjects. The inset images are provided for better visualization of local RGVP regions.

### 3.1 Comparison to state-of-the-art methods

We compare DeepRGVP to three SOTAs, including DeepWMA [19], DCNN [20] and SupWMA [24]. The SOTAs were designed to classify streamlines into different categories for tractography parcellation in the white matter of the brain. DCNN and DeepWMA use CNNs and streamline spatial coordinate features, SupWMA uses a point-cloud-based network with supervised contrastive learning, and DeepRGVP extends SupWMA to include additional microstructure information. We perform experimental comparison using training data with and without the proposed StreamDA. For the compared methods, we apply the suggested parameter settings described in the papers [19,20,24] and the authors' implementation.

Fig. 2 gives a visual comparison of the RGVP identified from each method using their best performing model. DCNN and DeepWMA are visually overinclusive with more streamlines compared to the ground truth (GT) (e.g. the local RGVP regions shown in the inset images of Subjects 1 and 2). DeepRGVP and SupWMA are visually similar to GT. But DeepRGVP has improved sensitivity in identifying local structures, e.g., the ipsilateral pathway passing through the optic chiasm in Subject 2 (indicated by the red arrows).

### 3.2 Ablation study

We performed an ablation study to evaluate the effects of MicroSCL and StreamDA, with comparison to the followings: 1) *baseline* that does not perform SCL (no supervised contrastive loss in our network) nor any DA; 2) $SCL_{label}$ that performs label-based SCL on original training data; 3) $SCL_{label+micro}$ that performs MicroSCL on original data; 4) $SCL_{label+micro}+Aug_{repetition}$ that performs MicroSCL with DA by simple duplication of RGVP streamlines; 5) $SCL_{label+micro}+Aug_{subsampling}$ that performs MicroSCL with StreamDA. Table 2 gives the comparison result, showing the proposed method generates the best performance.

Table 1. Quantitative comparisons with SOTA.

| Method | Data | Acc | F1 | Prec | Rec |
|---|---|---|---|---|---|
| *DeepWMA* | Original | 92.188 | 82.980 | 79.019 | **89.270** |
| *DCNN* | | 95.687 | 79.913 | 81.099 | 78.762 |
| *SupWMA* | | 96.117 | 82.149 | **82.281** | 82.017 |
| *Proposed* | | **96.268** | **83.070** | 82.117 | 84.045 |
| *DeepWMA* | StreamDA augmented | 93.449 | 84.614 | 81.957 | 88.034 |
| *DCNN* | | 93.141 | 73.789 | 63.204 | **88.634** |
| *SupWMA* | | 96.181 | 82.824 | 81.189 | 84.525 |
| *DeepRGVP* | | **96.646** | **84.316** | **85.928** | 82.764 |

As shown in Table 1, DeepRGVP generates the highest accuracy and the highest F1 using both original and augmented data, demonstrating the advantage of the proposed MicorSCL process. All methods except for DCNN achieve better performance using the augmented data compared to the original data, suggesting the benefit of our proposed DA strategy. (The decreased performance of DCNN when using the augmented (balanced) data may relate to the design of the DCNN method to perform well on unbalanced datasets [20].) In addition, DeepRGVP generates relatively high scores for both the precision and recall, demonstrating balanced classification performance between RGVP and non-RGVP streamlines.

Table 2. Ablation study results.

| Method | Acc | F1 | Prec | Rec |
|---|---|---|---|---|
| *Baseline* | 96.076 | 82.890 | 78.849 | **87.247** |
| $SCL_{label}$ | 96.117 | 82.149 | 82.281 | 82.017 |
| $SCL_{label+micro}$ | 96.268 | 83.070 | 82.117 | 84.045 |
| $SCL_{label+micro}+Aug_{repetition}$ | 96.449 | 83.410 | 84.909 | 81.964 |
| $SCL_{label+micro}+Aug_{subsampling}$ | **96.646** | **84.316** | **85.928** | 82.764 |

### 4. CONCLUSION

We present a novel microstructure-informed deep learning framework to enable automated identification of the RGVP. We propose a streamline-level data augmentation strategy for imbalanced tractography training data. Comparisons to several SOTA methods demonstrate DeepRGVP's improved performance. Future work could include applying DeepRGVP to data from different populations, e.g. patients with lesions affecting the RGVP. Overall, our study shows the high potential of using deep learning to automatically identify the RGVP.

## 5. COMPLIANCE WITH ETHICAL STANDARDS

This study was conducted retrospectively using public HCP imaging data [22]. No ethical approval was required.

## 6. ACKNOWLEDGEMENTS

We acknowledge the following NIH grants: P41EB015902, R01MH074794, R01MH125860 and R01MH119222. F.Z. acknowledges a BWH Radiology Research Pilot Grant Award.